\documentclass[11pt,a4paper]{article}
\usepackage[hyperref]{acl2018}
\usepackage{times}
\usepackage{latexsym}
\usepackage{graphicx}
\usepackage{subcaption}
\usepackage{booktabs}
\usepackage{array}
\usepackage{amsmath}
\usepackage{amssymb}
\usepackage{bbm}

\usepackage{url}

\aclfinalcopy 
\def\aclpaperid{1238} 

\graphicspath{ {imgs/} }

\newif\ifcomment
\commentfalse

\ifcomment
\newcommand{\uk}[1]{\textcolor{red}{UK: #1}}
\newcommand{\dan}[1]{\textcolor{green}{DJ: #1}}
\newcommand{\hh}[1]{\textcolor{blue}{HH: #1}}
\newcommand{\qp}[1]{\textcolor{cyan}{PQ: #1}}
\else
\newcommand{\uk}[1]{}
\newcommand{\dan}[1]{}
\newcommand{\hh}[1]{}
\newcommand{\qp}[1]{}
\fi

\newcommand{\nll}{\textrm{NLL}}
\newcommand{\pp}{\textrm{PP}}

\title{Sharp Nearby, Fuzzy Far Away: How Neural\\Language Models Use Context}

\author{Urvashi Khandelwal, He He, Peng Qi, Dan Jurafsky \\
  Computer Science Department \\
  Stanford University \\
  {\tt \{urvashik,hehe,pengqi,jurafsky\}@stanford.edu} \\}

\date{}

\begin{document}
\maketitle
\begin{abstract}
    We know very little about how neural language models (LM) use prior linguistic context. In this paper, we investigate the role of context in an LSTM LM, through ablation studies. Specifically, we analyze the increase in perplexity when prior context words are shuffled, replaced, or dropped. On two standard datasets, Penn Treebank and WikiText-2, we find that the model is capable of using about 200 tokens of context on average, but sharply distinguishes nearby context (recent 50 tokens) from the distant history. The model is highly sensitive to the order of words within the most recent sentence, but ignores word order in the long-range context (beyond 50 tokens), suggesting the distant past is modeled only as a rough semantic field or topic. We further find that the neural caching model~\cite{grave17cache} especially helps the LSTM to copy words from within this distant context. Overall, our analysis not only provides a better understanding of how neural LMs use their context, but also sheds light on recent success from cache-based models.

\end{abstract}


\section{Introduction}
\label{sect:intro}

Language models are an important component of natural language generation tasks, such as machine translation and summarization.
They use context (a sequence of words) to estimate a probability distribution of the upcoming word.
For several years now, neural language models (NLMs)~\cite{graves13generating,jozefowicz16exploring,grave17unbounded,dauphin17language,melis18lmeval,yang18breaking} have consistently outperformed classical $n$-gram models, an improvement often
attributed to their ability to model long-range dependencies in faraway context.
Yet, how these NLMs use the context is largely unexplained. 



Recent studies have begun to shed light on the information encoded by Long Short-Term Memory (LSTM) networks.
They can remember sentence lengths, word identity, and word order \cite{adi16fine},
can capture some syntactic structures such as subject-verb agreement \cite{linzen16assessing}, and can model
certain kinds of semantic compositionality such as negation and intensification \cite{li16visualizing}.




However, all of the previous work studies LSTMs at the sentence level, even though they can potentially encode longer context.
Our goal is to complement the prior work to provide a richer understanding of the role of context, in particular, long-range context beyond a sentence.
We aim to answer the following questions:
(i) How much context is used by NLMs, in terms of the number of tokens?
(ii) Within this range, are nearby and long-range contexts represented differently?
(iii) How do copy mechanisms help the model use different regions of context?

We investigate these questions via ablation studies on a standard LSTM language model~\cite{merity18regopt}
on two benchmark language modeling datasets: Penn Treebank and WikiText-2.
Given a pretrained language model,
we perturb the prior context in various ways \emph{at test time}, to study how much the  perturbed information affects model performance.
Specifically,
we alter the context length to study how many tokens are used,
permute tokens to see if LSTMs care about word order in both local and global contexts,
and drop and replace target words to test the copying abilities of LSTMs with and without an external copy mechanism, such as the neural cache
~\cite{grave17cache}.
The cache operates by first recording target words and their context representations seen in the history, and then encouraging the model to copy a word from the past when the current context representation matches that word's recorded context vector.



We find that the LSTM is capable of using about 200 tokens of context on average,
with no observable differences from changing the hyperparameter settings. 
Within this context range, word order is only relevant within the 20 most recent tokens or about a sentence. 
In the long-range context,
order has almost no effect on performance, suggesting that the model maintains a high-level, rough semantic representation of faraway words.
Finally, we find that LSTMs can regenerate some words seen in the nearby context,
but heavily rely on the cache to help them copy words from the long-range context.

\section{Language Modeling}
\label{sect:lm}

Language models assign probabilities to sequences of words.
In practice, the probability can be factorized using the chain rule
$$P(w_1,\ldots,w_{t}) = \prod_{i=1}^{t}P(w_i|w_{i-1},\ldots,w_{1}),$$
and language models compute the conditional probability of a \emph{target word} $w_t$ given its preceding context, $w_1, \ldots, w_{t-1}$.


Language models are trained to minimize the negative log likelihood of the training corpus:
\[\nll = -\frac{1}{T}\sum_{t=1}^{T}\log P(w_t|w_{t-1}, \ldots, w_1),\]
and the model's performance is usually evaluated
by perplexity ($\pp$) on a held-out set:
\[\pp=\exp(\nll).\]
When testing the effect of ablations, we focus on comparing differences in the language model's losses ($\nll$) on the dev set, which is equivalent to relative improvements in perplexity. 


\section{Approach}
\label{sect:approach}

\begin{table}[t]
	\centering
	\small
	\setlength{\tabcolsep}{0.4em}
	\begin{tabular}{m{2.75cm}cccc}
		\toprule
		& \multicolumn{2}{c}{\textbf{PTB}} & \multicolumn{2}{c}{\textbf{Wiki}} \\
		& Dev & Test & Dev & Test\\
		\midrule	
		\# Tokens &  73,760 & 82,430 & 217,646 & 245,569 \\
		Perplexity (no cache) & 59.07 & 56.89 & 67.29 & 64.51 \\
		Avg. Sent. Len. & 20.9 & 20.9 & 23.7 & 22.6 \\
		\bottomrule
	\end{tabular}
	\caption{Dataset statistics and performance relevant to our experiments.}
	\label{tab:data}
\end{table}

Our goal is to investigate the effect of contextual features such as the length of context, word order and more, on LSTM performance. Thus, we use ablation analysis, during evaluation, to measure changes in model performance in the absence of certain contextual information. 

Typically, when testing the language model on a held-out sequence of words,
all tokens prior to the target word are fed to the model;
we call this the \emph{infinite-context} setting.
In this study, we observe the change in perplexity or $\nll$
when the model is fed a perturbed context $\delta(w_{t-1},\ldots,w_1)$, at test time. $\delta$ refers to the perturbation function, and
we experiment with perturbations such as dropping tokens, shuffling/reversing tokens, and replacing tokens with other words from the vocabulary.\footnote{Code for our experiments available at \url{https://github.com/urvashik/lm-context-analysis}} It is important to note that we do not train the model with these perturbations. This is because the aim is to start with an LSTM that has been trained in the standard fashion, and discover how much context it uses and which features in nearby vs. long-range context are important. Hence, the mismatch in training and test is a necessary part of experiment design, and all measured losses are upper bounds which would likely be lower, were the model also trained to handle such perturbations.

We use a standard LSTM language model, trained and finetuned using the Averaging SGD optimizer \cite{merity18regopt}.\footnote{Public release of their code at \url{https://github.com/salesforce/awd-lstm-lm}}
We also augment the model with a cache \emph{only} for Section~\ref{sect:cache}, in order to investigate why an external copy mechanism is helpful.
A short description of the architecture and a detailed list of hyperparameters is listed in Appendix~\ref{app:hps}, and we refer the reader to the original paper for additional details.

We analyze two datasets commonly used for language modeling, Penn Treebank (PTB)~\cite{marcus93building,mikolov10ptb} and Wikitext-2 (Wiki)~\cite{merity17pointer}.
PTB consists of Wall Street Journal news articles with 0.9M tokens for training and a 10K vocabulary.
Wiki is a larger and more diverse dataset,
containing Wikipedia articles across many topics with 2.1M tokens for training and a 33K vocabulary.
Additional dataset statistics are provided in Table~\ref{tab:data}.

In this paper, we present results only on the dev sets, in order to avoid revealing details about the test sets. However, we have confirmed that all results are consistent with those on the test sets. In addition, for all experiments we report averaged results from three models trained with different random seeds. Some of the figures provided contain trends from only one of the two datasets and the corresponding figures for the other dataset are provided in Appendix~\ref{app:figs}.


\section{How much context is used?}
\label{sect:contsize}
\begin{figure*}[t!]
	\centering
	\begin{subfigure}[t]{0.49\textwidth}
		\centering
		\includegraphics[width=0.85\textwidth]{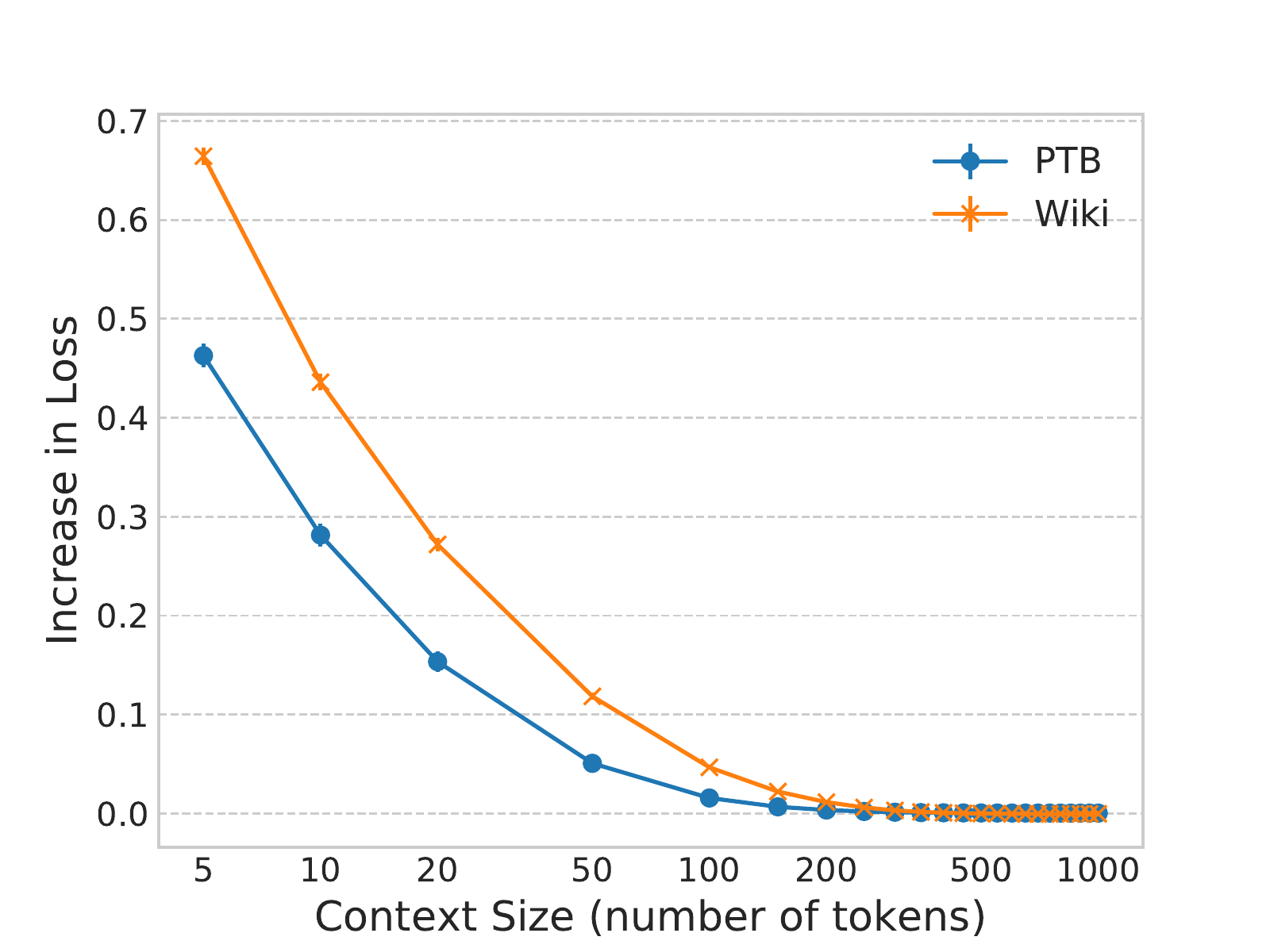}
		\caption{Varying context size.}		
		\label{fig:contextsizefull}
	\end{subfigure}
	\hfill
	\begin{subfigure}[t]{0.49\textwidth}
		\centering
		\includegraphics[width=0.85\textwidth]{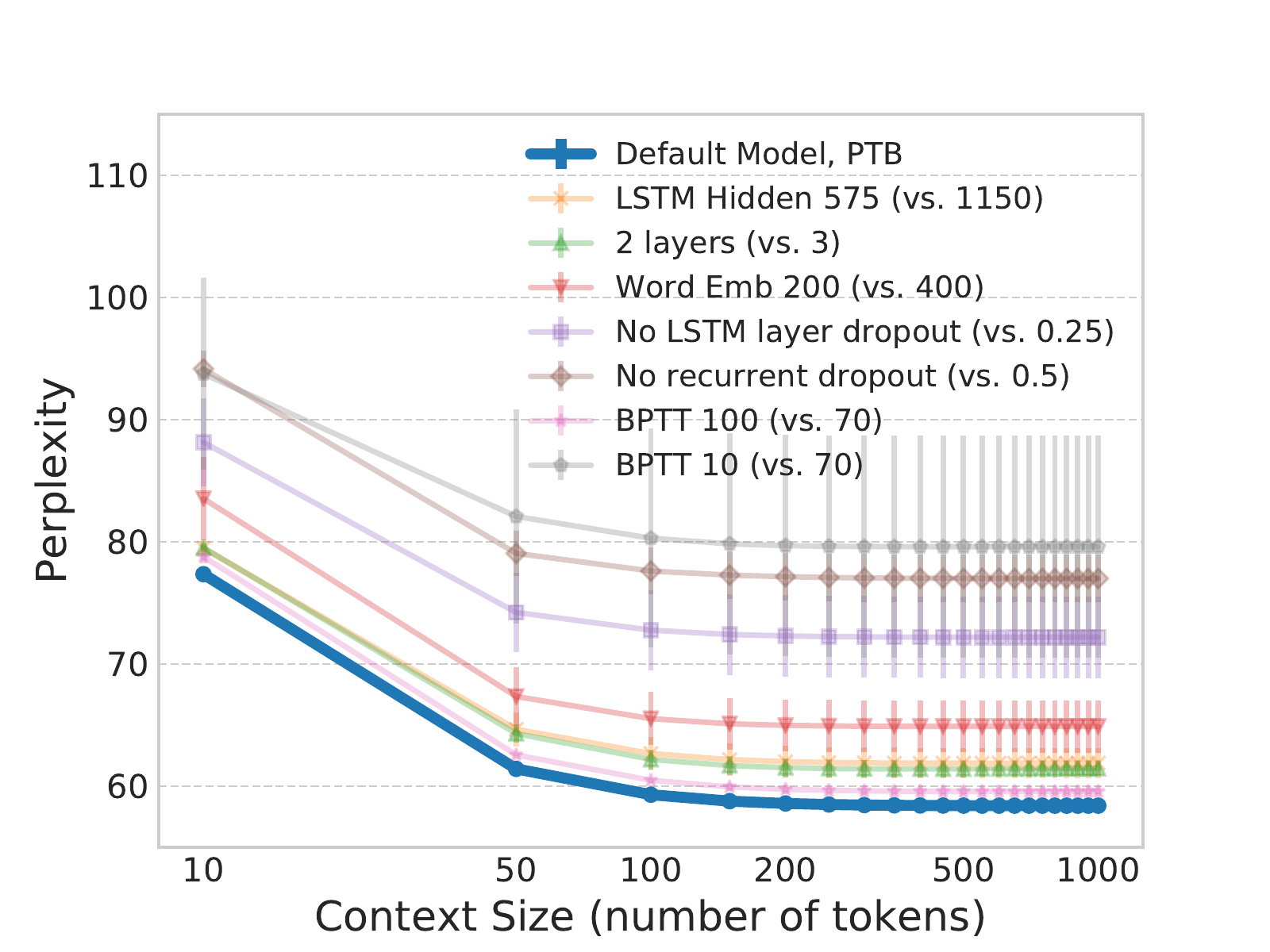}
		\caption{Changing model hyperparameters.}
		\label{fig:contexthps}
	\end{subfigure}
	\begin{subfigure}[t]{0.49\textwidth}
		\centering
		\includegraphics[width=0.85\textwidth]{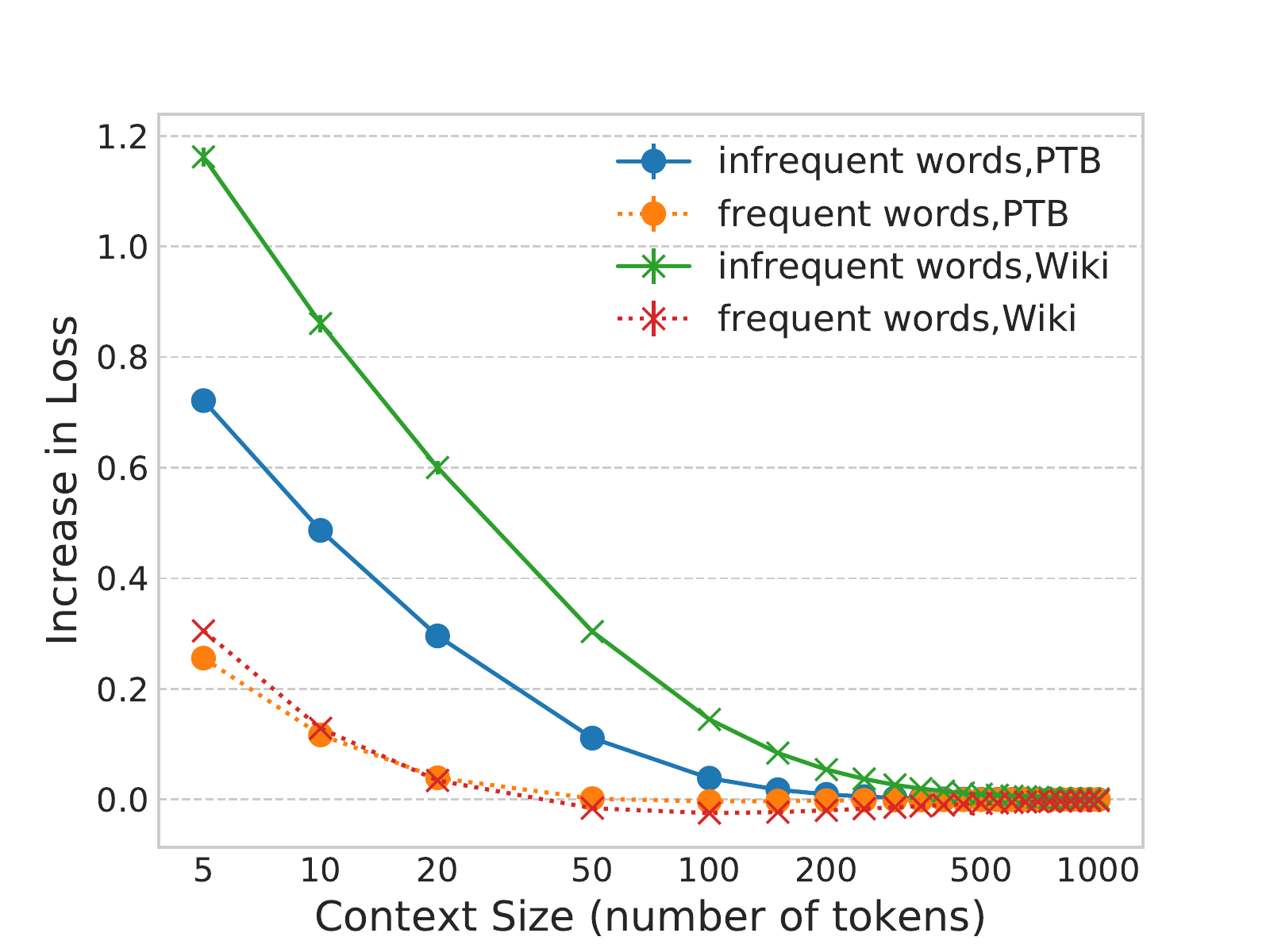}
		\caption{Frequent vs. infrequent words.}		
		\label{fig:freqcontextsize}
	\end{subfigure}
	\hfill
	\begin{subfigure}[t]{0.49\textwidth}
		\centering
		\includegraphics[width=0.85\textwidth]{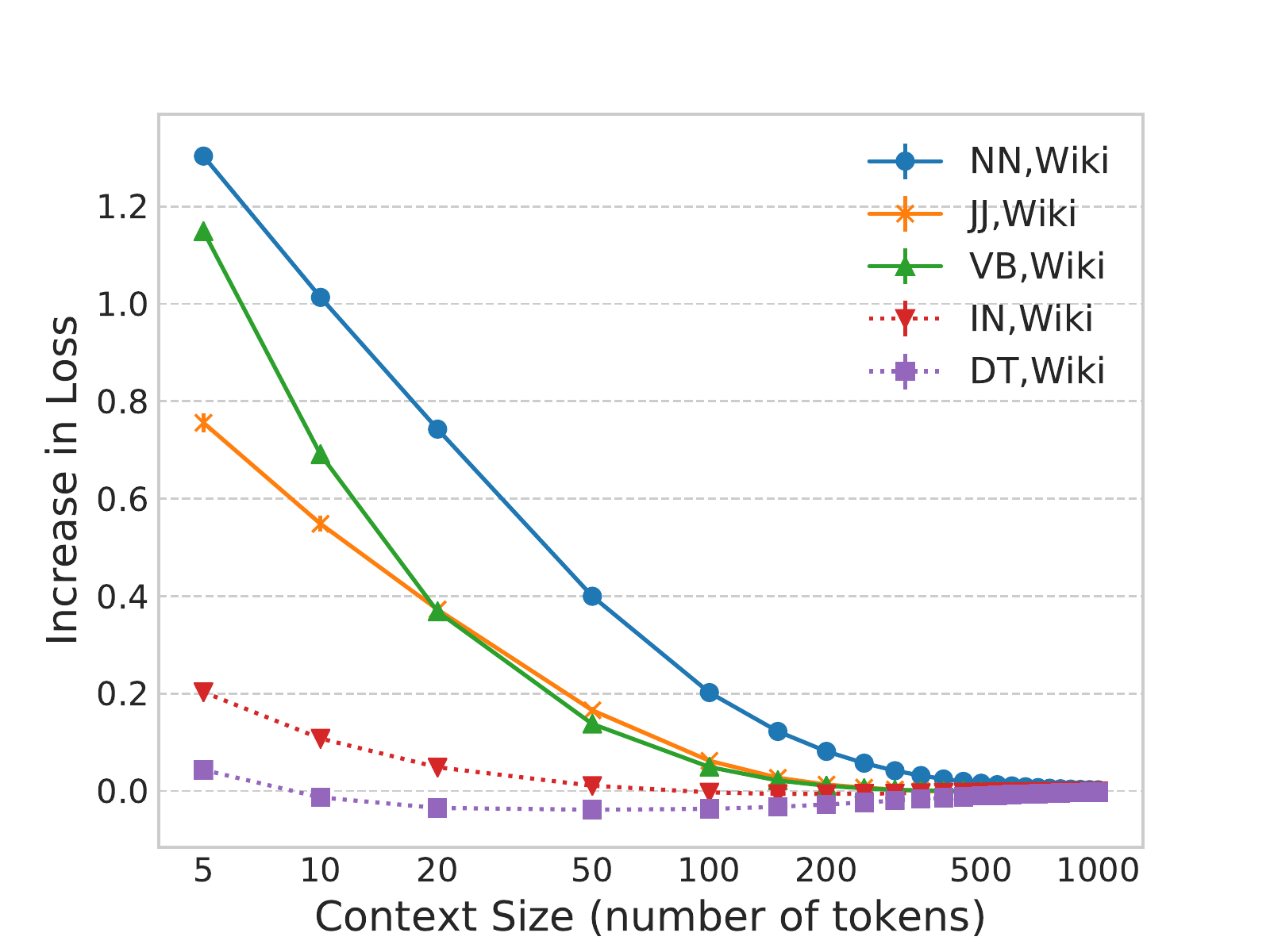}
		\caption{Different parts-of-speech.}		
		\label{fig:poscontextsize}
	\end{subfigure}
	\caption{Effects of varying the number of tokens provided in the context, as compared to the same model provided with infinite context. Increase in loss represents an absolute increase in $\nll$ over the entire corpus, due to restricted context. All curves are averaged over three random seeds, and error bars represent the standard deviation. \textbf{(a)} The model has an effective context size of 150  on PTB and 250 on Wiki. \textbf{(b)} Changing model hyperparameters does not change the context usage trend, but does change model performance. We report perplexities to highlight the consistent trend. \textbf{(c)} Infrequent words need more context than frequent words. \textbf{(d)} Content words need more context than function words.}
	\label{fig:contextsize}
\end{figure*}



LSTMs are designed to capture long-range dependencies in sequences~\cite{hochreiter97long}. In practice, LSTM language models are provided an infinite amount of prior context, which is as long as the test sequence goes. However, it is unclear how much of this history has a direct impact on model performance. In this section, we investigate how many tokens of context achieve a similar loss (or 1-2\% difference in model perplexity) to providing the model infinite context. We consider this the \emph{effective context size}.


\paragraph{LSTM language models have an effective context size of about 200 tokens on average.}
We determine the effective context size by varying the number of tokens fed to the model.
In particular, at test time, we feed the model the most recent $n$ tokens:
\begin{equation}
    \delta_{\textrm{truncate}}(w_{t-1},\ldots,w_{1}) = (w_{t-1},\ldots,w_{t-n}),
	\label{eq:contextsize}
\end{equation}
where $n > 0$ and all tokens farther away from the target $w_t$ are dropped.\footnote{Words at the beginning of the test sequence with fewer than $n$ tokens in the context are ignored for loss computation.}
We compare the dev loss ($\nll$) from truncated context,
to that of the infinite-context setting where all previous words are fed to the model.
The resulting increase in loss indicates how important the dropped tokens are for the model.

Figure~\ref{fig:contextsizefull} shows that the difference in dev loss,
between truncated- and infinite-context variants of the test setting, 
gradually diminishes as we increase $n$ from 5 tokens to 1000 tokens.
In particular, we only see a 1\% increase in perplexity as we move beyond a context of 150 tokens on PTB and 250 tokens on Wiki.
Hence, we provide empirical evidence to show that LSTM language models do, in fact, model long-range dependencies, without help from extra context vectors or caches.

\paragraph{Changing hyperparameters does not change the effective context size.}
NLM performance has been shown to be sensitive to hyperparameters such as the dropout rate and model size~\cite{melis18lmeval}.
To investigate if these hyperparameters affect the effective context size as well, we train separate models by varying the following hyperparameters one at a time:
(1) number of timesteps for truncated back-propogation
(2) dropout rate,
(3) model size (hidden state size, number of layers, and word embedding size).
In Figure~\ref{fig:contexthps}, we show that while different hyperparameter settings result in different perplexities in the infinite-context setting,
the trend of how perplexity changes as we reduce the context size remains the same.

\subsection{Do different types of words need different amounts of context?}
\label{sec:adaptive}

The effective context size determined in the previous section is aggregated over the entire corpus, which ignores the type of the upcoming word.
\citet{boyd09syntactic} have previously investigated the differences in context used by different types of words and found that function words rely on less context than content words. We investigate whether the effective context size varies across different types of words, by 
categorizing them based on either frequency or parts-of-speech.
Specifically, we vary the number of context tokens in the same way as the previous section, and aggregate loss over words within each class separately.

\paragraph{Infrequent words need more context than frequent words.} We categorize words that appear at least 800 times in the training set as \emph{frequent}, and the rest as \emph{infrequent}. Figure~\ref{fig:freqcontextsize} shows that the loss of frequent words is insensitive to missing context beyond the 50 most recent tokens, which holds across the two datasets. Infrequent words, on the other hand, require more than 200 tokens. 

\paragraph{Content words need more context than function words.}
Given the parts-of-speech of each word,
we define \emph{content words} as nouns, verbs and adjectives, and \emph{function words} as prepositions and determiners.%
\footnote{We obtain part-of-speech tags using Stanford CoreNLP~\cite{manning14stanford}.}
Figure~\ref{fig:poscontextsize} shows that the loss of nouns and verbs is affected by distant context,
whereas when the target word is a determiner, the model only relies on words within the last 10 tokens.

\paragraph{Discussion.} Overall, we find that the model's effective context size is dynamic. It depends on the target word,
which is consistent with what we know about language, e.g., determiners require less context than nouns \cite{boyd09syntactic}. In addition, these findings are consistent with those previously  reported for different language models and datasets \cite{hill15goldilocks,wang16largecontext}.



\section{Nearby vs. long-range context}
\label{sect:order}

\begin{figure*}
	\centering
	\begin{subfigure}[t]{0.49\textwidth}
		\centering
		\includegraphics[width=\textwidth]{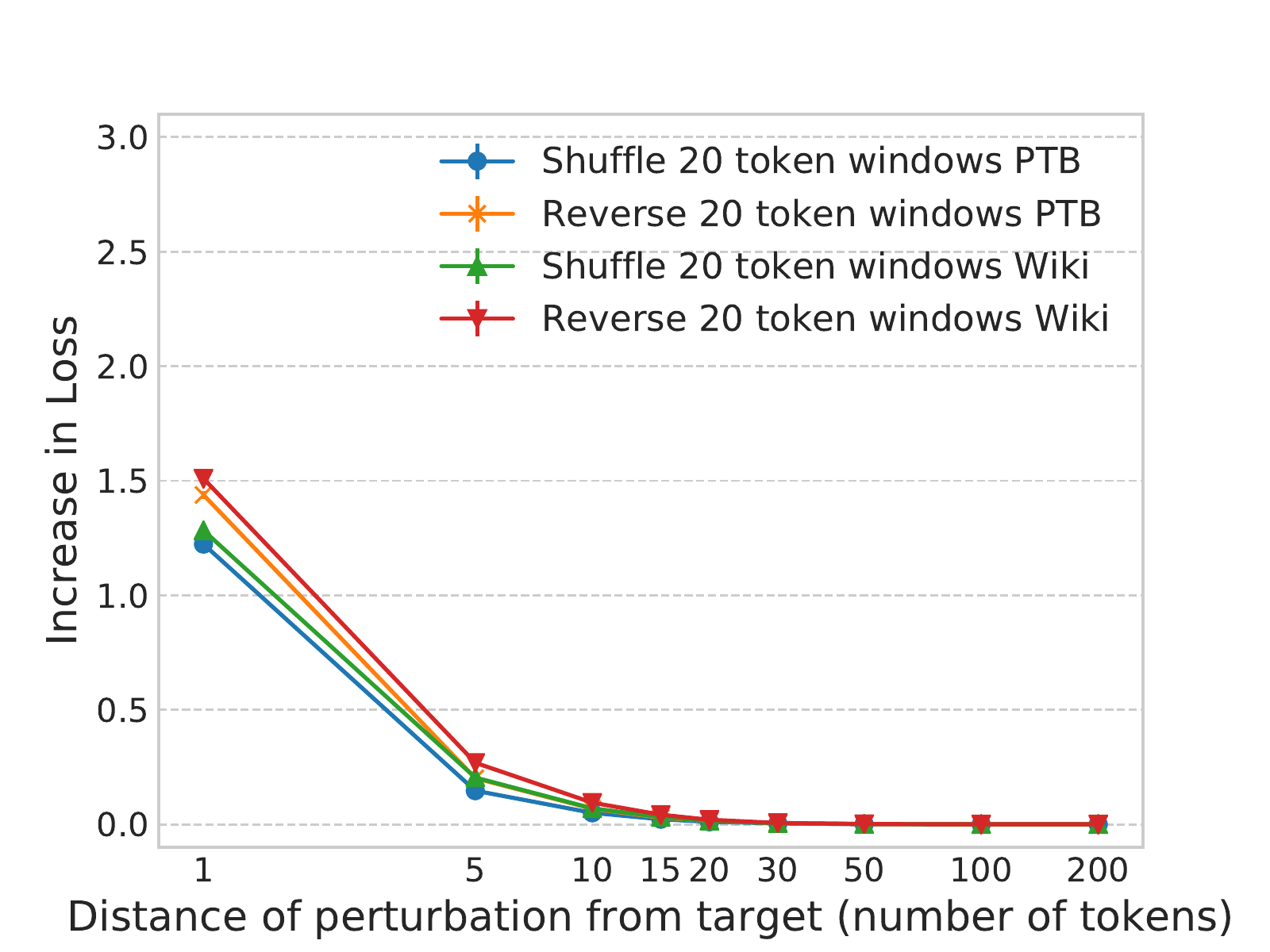}
		\caption{Perturb order locally, within 20 tokens of each point.}
		\label{fig:localorder}
	\end{subfigure}
	\hfill
	\begin{subfigure}[t]{0.49\textwidth}
		\centering
		\includegraphics[width=\textwidth]{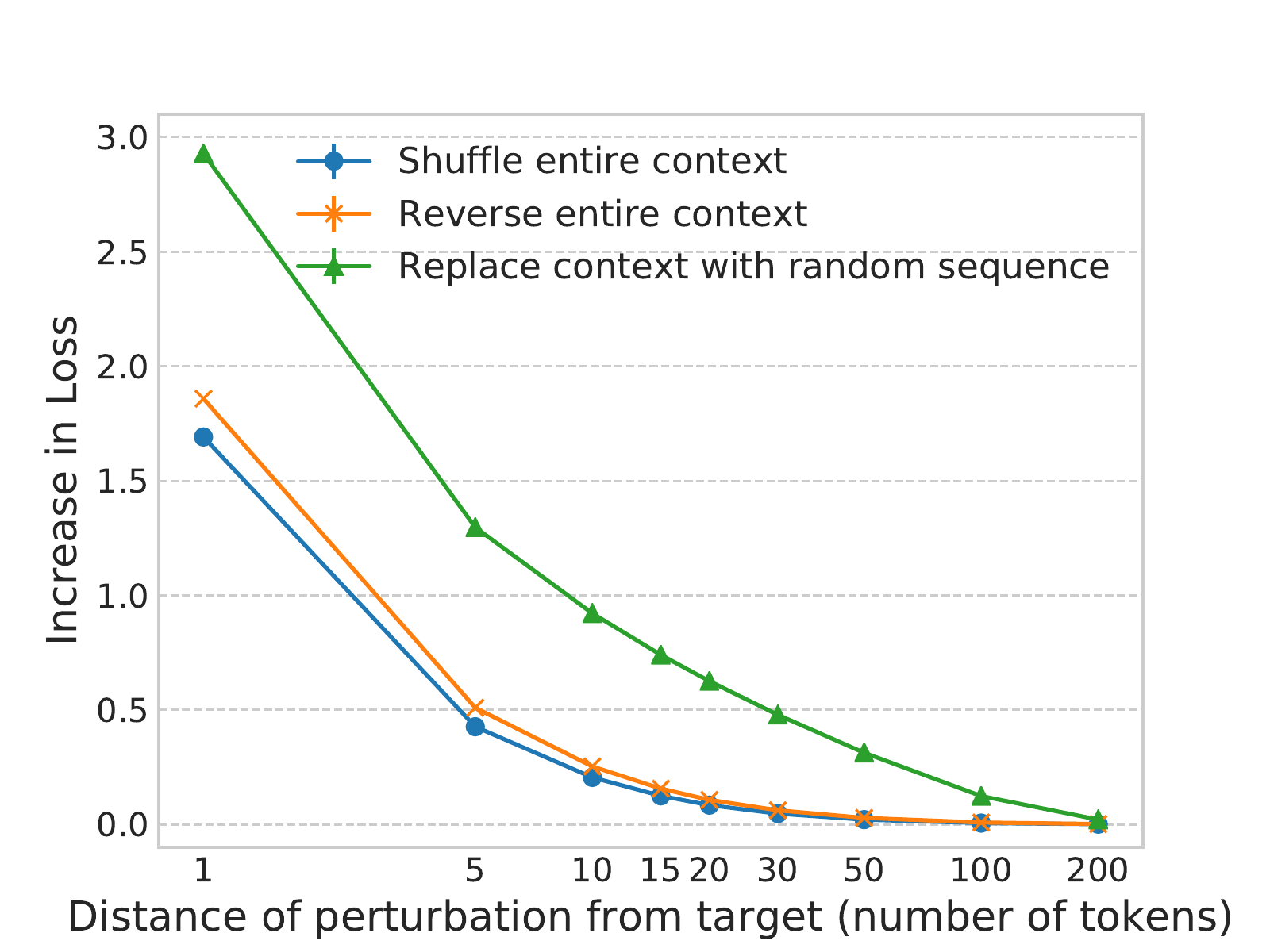}
		\caption{\centering{Perturb global order, i.e. all tokens in the context before a given point, in Wiki.}}
		\label{fig:globalorder}
	\end{subfigure}
	\caption{Effects of shuffling and reversing the order of words in 300 tokens of context, relative to an unperturbed baseline. All curves are averages from three random seeds, where error bars represent the standard deviation. \textbf{(a)} Changing the order of words within a 20-token window has negligible effect on the loss after the first 20 tokens. \textbf{(b)} Changing the global order of words within the context does not affect loss beyond 50 tokens.}
	\label{fig:order}
\end{figure*}

An effective context size of 200 tokens allows for representing linguistic information at many levels of abstraction, such as words, sentences, topics, etc. 
In this section, we investigate the importance of contextual information such as word order and word identity.
Unlike prior work that studies LSTM embeddings at the sentence level,
we look at both nearby and faraway context,
and analyze how the language model treats contextual information presented in different regions of the context.

\subsection{Does word order matter?}
\label{sect:wordorder} 
\citet{adi16fine} have shown that LSTMs are aware of word order within a sentence. We investigate whether LSTM language models are sensitive to word order within a larger context window.
To determine the range in which word order affects model performance, we permute substrings in the context to observe their effect on dev loss compared to the unperturbed baseline.
In particular, we perturb the context as follows,
\begin{equation}
\begin{split}
    &\delta_{\textrm{permute}}(w_{t-1},\ldots,w_{t-n}) = \\
    &(w_{t-1},..,\rho(w_{t-s_1-1},.., w_{t-s_2}),..,w_{t-n})
\end{split}
\end{equation}
where $\rho \in \{\mathrm{shuffle}, \mathrm{reverse}\}$ and $(s_1, s_2]$ 
denotes the range of the substring to be permuted. We refer to this substring as the \emph{permutable span}.
For the following analysis, we distinguish \emph{local word order}, within 20-token permutable spans which are the length of an average sentence,
from \emph{global word order}, which extends beyond local spans to include all the farthest tokens in the history.
We consider selecting permutable spans within a context of $n=300$ tokens,
which is greater than the effective context size.

\paragraph{Local word order only matters for the most recent 20 tokens.}


We can locate the region of context beyond which the local word order has no relevance, by permuting word order locally at various points within the context. 
We accomplish this by varying $s_1$ and setting $s_2=s_1+20$.
Figure~\ref{fig:localorder} shows that local word order matters very much within the most recent 20 tokens, and far less beyond that.

\paragraph{Global order of words only matters for the most recent 50 tokens.}
Similar to the local word order experiment, we locate the point beyond which the general location of words within the context is irrelevant, by permuting global word order.
We achieve this by varying $s_1$ and fixing $s_2=n$.
Figure~\ref{fig:globalorder} demonstrates that after 50 tokens, shuffling or reversing the remaining words in the context has no effect on the model performance.

In order to determine whether this is due to insensitivity to word order
or whether the language model is simply not sensitive to any changes in the long-range context,
we further replace words in the permutable span with a randomly sampled sequence of the same length from the training set. The gap between the permutation and replacement curves in Figure~\ref{fig:globalorder} illustrates that the identity of words in the far away context is still relevant, and only the order of the words is not.

\paragraph{Discussion.} These results suggest that word order matters only within the most recent sentence, beyond which the order of sentences matters for 2-3 sentences (determined by our experiments on global word order). After 50 tokens, word order has almost no effect, but the identity of those words is still relevant, suggesting a high-level, rough semantic representation for these faraway words. In light of these observations, we define 50 tokens as the boundary between nearby and long-range context, for the rest of this study. Next, we investigate the importance of different word types in the different regions of context.

\begin{figure}[h]
	\includegraphics[width=\linewidth]{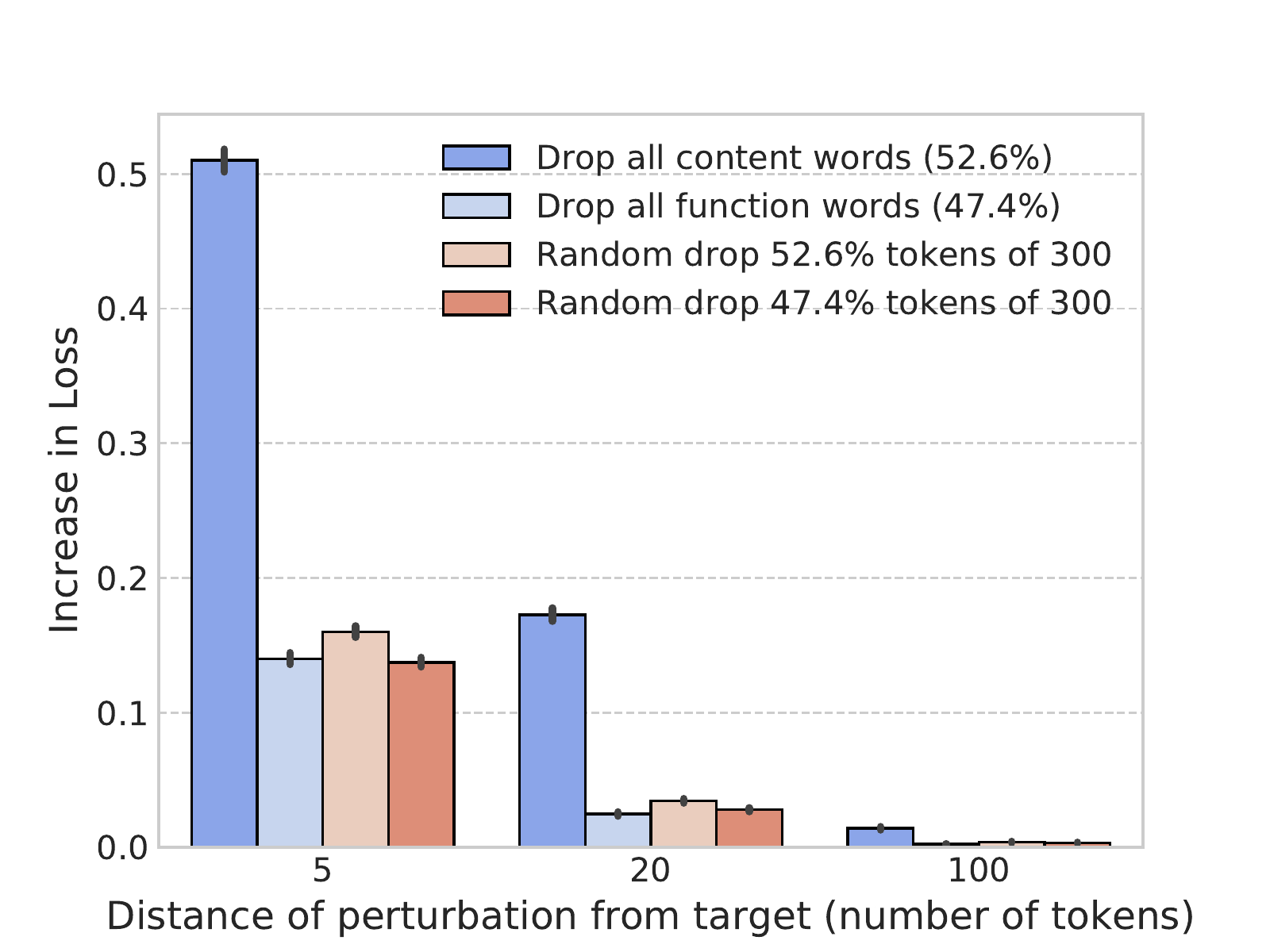}
	\caption{Effect of dropping content and function words from 300 tokens of context relative to an unperturbed baseline, on PTB. Error bars represent 95\% confidence intervals. Dropping both content and function words 5 tokens away from the target results in a nontrivial increase in loss, whereas beyond 20 tokens, only content words are relevant.}
	\label{fig:posdrop}
\end{figure}
\subsection{Types of words and the region of context}
\label{sect:wordtypes}
Open-class or \emph{content words} such as nouns, verbs, adjectives and adverbs, contribute more to the semantic context of natural language than \emph{function words} such as determiners and prepositions.
Given our observation that the language model represents long-range context as a rough semantic representation, a natural question to ask is
how important are function words in the long-range context? 
Below, we study the effect of these two classes of words on the model's performance. Function words are defined as all words that are not nouns, verbs, adjectives or adverbs.

\paragraph{Content words matter more than function words.}
To study the effect of content and function words on model perplexity,
we drop them from different regions of the context and compare the resulting change in loss.
Specifically, we perturb the context as follows,
\begin{equation}
\label{eq:drop}
\begin{split}
    &\delta_{\textrm{drop}}(w_{t-1},\ldots,w_{t-n}) =\\
    &(w_{t-1},..,w_{t-s_1},f_{\mathrm{pos}}(y, (w_{t-s_1-1},..,w_{t-n})))
\end{split}
\end{equation}
where $f_{\mathrm{pos}}(y, \textrm{span})$ is a function that drops all words with POS tag $y$ in a given span.
$s_1$ denotes the starting offset of the perturbed subsequence.
For these experiments, we set $s_1 \in \{5,20,100\}$.
On average, there are slightly more content words than function words in any given text.
As shown in Section~\ref{sect:contsize}, dropping more words results in higher loss.
To eliminate the effect of dropping different fractions of words,
for each experiment where we drop a specific word type,
we add a control experiment where 
the same number of tokens are sampled randomly from the context, and dropped.

Figure~\ref{fig:posdrop} shows that dropping content words as close as 5 tokens from the target word increases model perplexity by about 65\%, whereas dropping the same proportion of tokens at random, results in a much smaller 17\% increase. Dropping all function words, on the other hand, is not very different from dropping the same proportion of words at random, but still increases loss by about 15\%. This suggests that within the most recent sentence, content words are extremely important but function words are also relevant since they help maintain grammaticality and syntactic structure. On the other hand, beyond a sentence, only content words have a sizeable influence on model performance.


\section{To cache or not to cache?}
\label{sect:content}

As shown in Section~\ref{sect:wordorder}, LSTM language models use a high-level, rough semantic representation for long-range context,  suggesting that they might not be using information from any specific words located far away.
\citet{adi16fine} have also shown that while LSTMs are aware of which words appear in their context, this awareness degrades with increasing length of the sequence. 
However, the success of copy mechanisms such as attention and caching~\cite{bahdanau14attn,hill15goldilocks,merity17pointer,grave17unbounded,grave17cache} suggests that information in the distant context is very useful.
Given this fact, can LSTMs copy any words from  context without relying on external copy mechanisms? Do they copy words from nearby and long-range context equally?
How does the caching model help?
In this section, we investigate these questions by studying how LSTMs copy words from different regions of context. More specifically, we look at two regions of context, nearby (within 50 most recent tokens) and long-range (beyond 50 tokens), and study three categories of target words: those that can be copied from nearby context ($C_{\text{near}}$), those that can \emph{only} be copied from long-range context ($C_{\text{far}}$), and those that cannot be copied at all given a limited context ($C_{\text{none}}$).

\subsection{Can LSTMs copy words without caches?}
\label{sect:copy}

\begin{figure*}
	\centering
	\begin{subfigure}[t]{0.49\textwidth}
		\centering
		\includegraphics[width=0.9\textwidth]{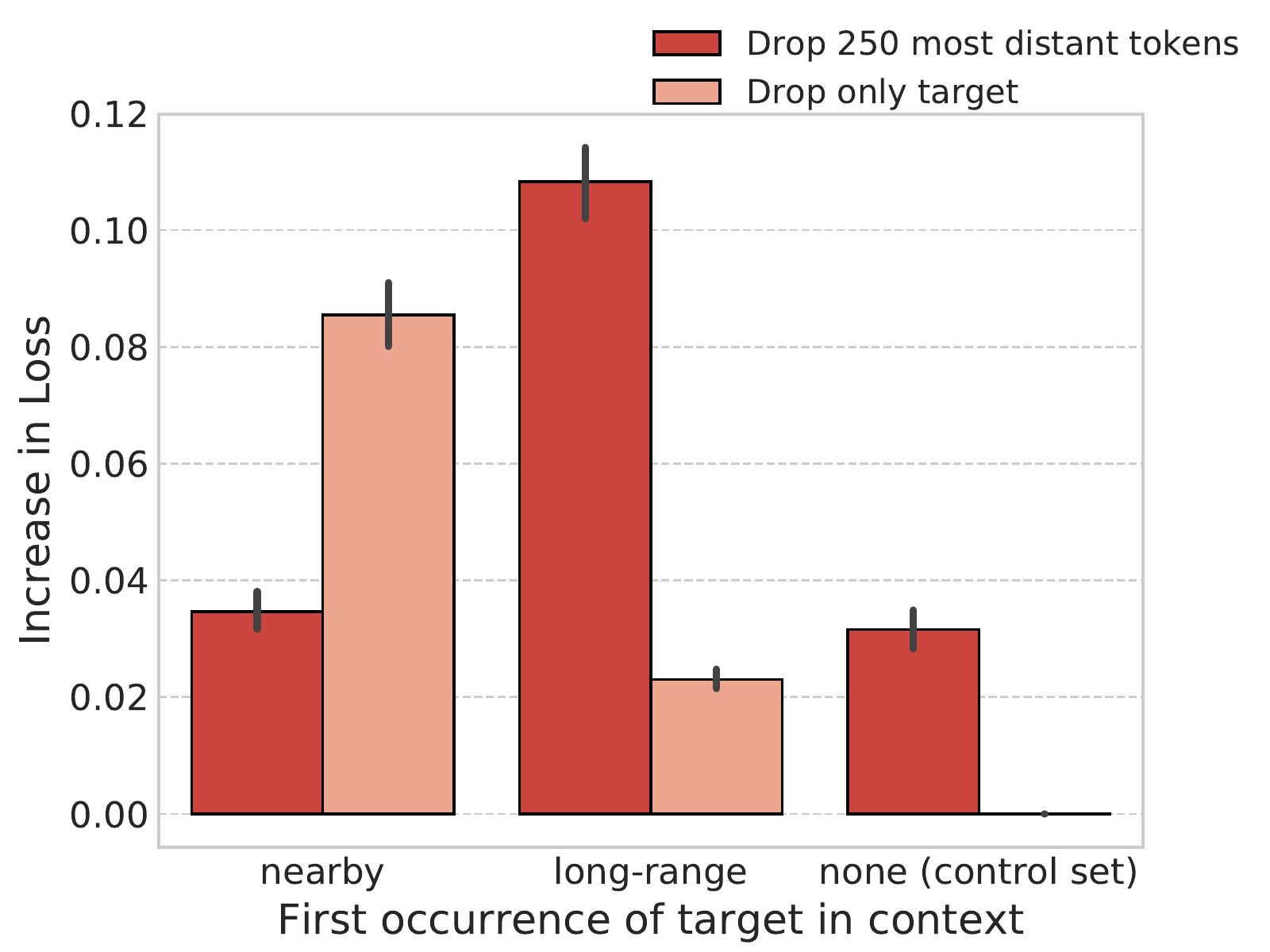}
		\caption{Dropping tokens}
		\label{fig:droptarget}
	\end{subfigure}
	\hfill
	\begin{subfigure}[t]{0.49\textwidth}
		\centering
		\includegraphics[width=0.9\textwidth]{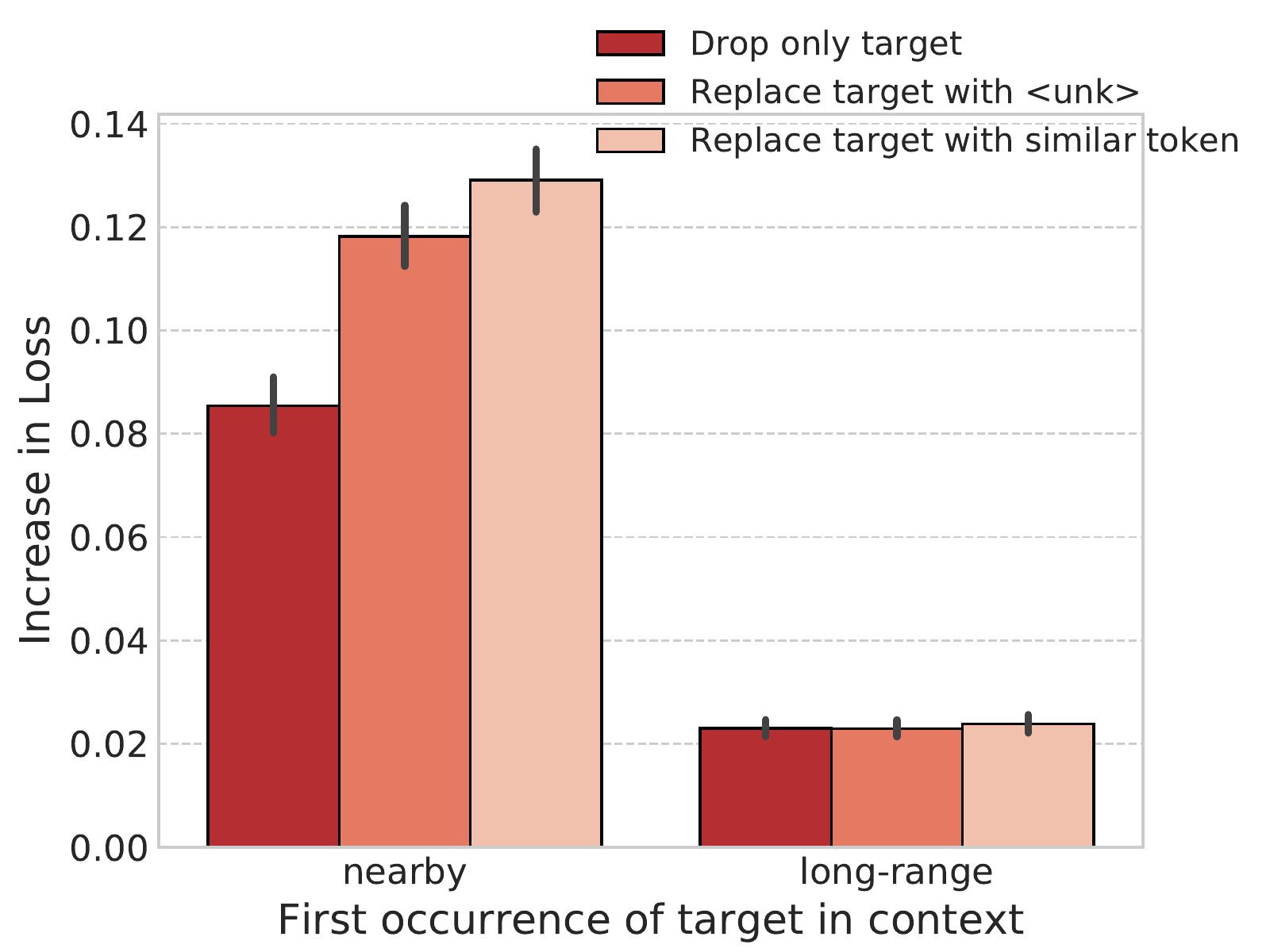}
		\caption{Perturbing occurrences of target word in context.}
		\label{fig:perturbtarget}
	\end{subfigure}
	\caption{Effects of perturbing the target word in the context compared to dropping long-range context altogether, on PTB. Error bars represent 95\% confidence intervals. \textbf{(a)} Words that can only be copied from long-range context are more sensitive to dropping all the distant words than to dropping the target. For words that can be copied from nearby context, dropping only the target has a much larger effect on loss compared to dropping the long-range context. \textbf{(b)} Replacing the target word with other tokens from vocabulary hurts more than dropping it from the context, for words that can be copied from nearby context, but has no effect on words that can only be copied from far away.}
	\label{fig:target}
\end{figure*}
Even without a cache, LSTMs often regenerate words that have already appeared in prior context. We investigate how much the model relies on the previous occurrences of the upcoming target word, by analyzing the change in loss after dropping and replacing this target word in the context.

\paragraph{LSTMs can regenerate words seen in nearby context.} 
In order to demonstrate the usefulness of target word occurrences in context, we experiment with dropping all the distant context versus dropping only occurrences of the target word from the context.
In particular, we compare removing all tokens after the 50 most recent tokens, (Equation~\ref{eq:contextsize} with $n=50$), versus removing only the target word, in context of size $n=300$:
\begin{equation}
	\label{eq:droptarget}
	\begin{split}
        \delta_{\textrm{drop}}(w_{t-1}&,\ldots,w_{t-n}) =\\ &f_{\textrm{word}}(w_t, (w_{t-1},\ldots,w_{t-n})),
	\end{split}
\end{equation}
where $f_{\textrm{word}}(w, \textrm{span})$ drops words equal to $w$ in a given span.
We compare applying both perturbations to a baseline model with unperturbed context restricted to $n=300$. 
We also include the target words that never appear in the context ($C_{\text{none}}$) as a control set for this experiment.

The results show that LSTMs rely on the rough semantic representation of the faraway context to generate $C_{\text{far}}$,
but direclty copy $C_{\text{near}}$ from the nearby context.
In Figure~\ref{fig:droptarget}, the long-range context bars show that
for words that can only be copied from long-range context ($C_{\text{far}}$), removing all distant context 
is far more disruptive than removing only occurrences of the target word
(12\% and 2\% increase in perplexity, respectively).
This suggests that the model relies more on the rough semantic representation of faraway context
to predict these $C_{\text{far}}$ tokens,
rather than directly copying them from the distant context.
On the other hand, for words that can be copied from nearby context ($C_{\text{near}}$), removing all long-range context has a smaller effect (about 3.5\% increase in perplexity) as seen in Figure~\ref{fig:droptarget}, compared to removing the target word which increases perplexity by almost 9\%.
This suggests that these $C_{\text{near}}$ tokens are more often copied from nearby context, than inferred from information found in the rough semantic representation of long-range context.

\begin{figure*}
	\includegraphics[width=\linewidth]{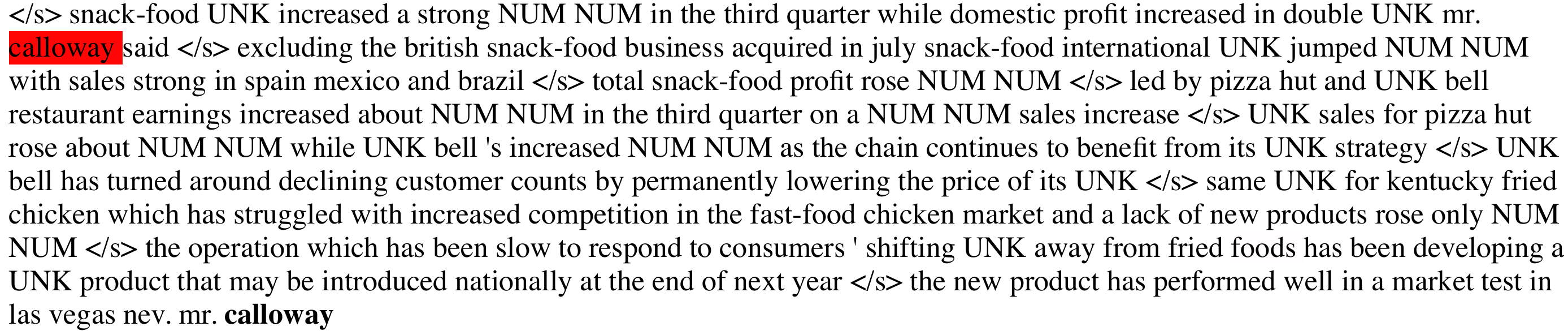}
	\caption{Success of neural cache on PTB. Brightly shaded region shows peaky distribution.}
	\label{fig:ptb_good_case}
\end{figure*}
\begin{figure*}
	\includegraphics[width=\textwidth]{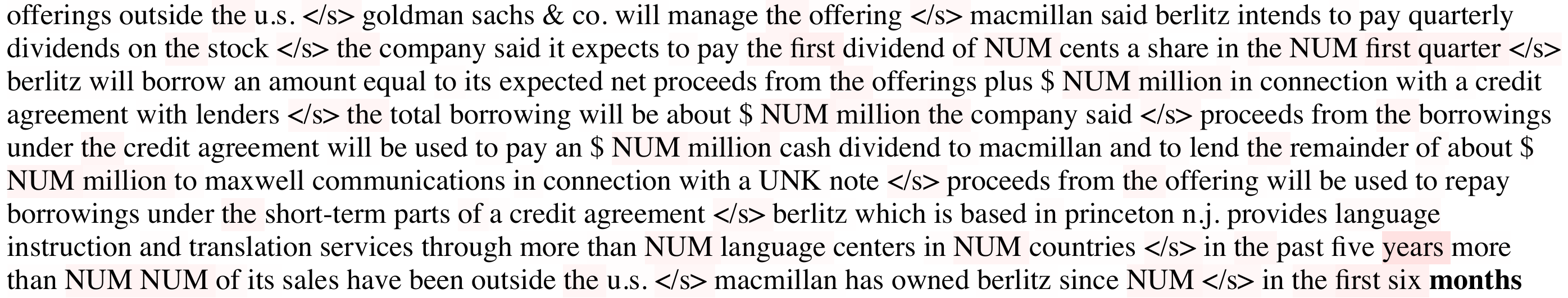}
	\caption{Failure of neural cache on PTB. Lightly shaded regions show flat distribution.}
	\label{fig:ptb_bad_case}
\end{figure*}

However, is it possible that dropping the target tokens altogether, hurts the model too much by adversely affecting grammaticality of the context?
We test this theory by replacing target words in the context with other words from the vocabulary. This perturbation is similar to Equation~\ref{eq:droptarget}, except instead of dropping the token, we replace it with a different one.
In particular, we experiment with replacing the target with \verb|<unk>|, to see if having the generic word is better than not having any word. We also replace it with a word that has the same part-of-speech tag and a similar frequency in the dataset, to observe how much this change confuses the model. Figure~\ref{fig:perturbtarget} shows that replacing the target with other words results in up to a 14\% increase in perplexity for $C_{\text{near}}$, which suggests that the replacement token seems to confuse the model far more than when the token is simply dropped. However, the words that rely on the long-range context, $C_{\text{far}}$, are largely unaffected by these changes,
which confirms our conclusion from dropping the target tokens:
$C_{\text{far}}$ words are predicted from the rough representation of faraway context
instead of specific occurrences of certain words.

\subsection{How does the cache help?}
\label{sect:cache}
If LSTMs can already regenerate words from nearby context, how are copy mechanisms helping the model?
We answer this question by analyzing how the neural cache model \cite{grave17cache} helps with improving model performance. 
The cache records the hidden state $h_t$ at each timestep $t$, and computes a cache distribution over the words in the history as follows:
\begin{equation}
\begin{split}
P_{\mathrm{cache}}&(w_{t}| w_{t-1},\ldots,w_{1};h_t,\ldots,h_{1})\\&\propto \sum_{i=1}^{t-1}{\mathbbm{1}[w_i = w_{t}]\exp(\theta h_i^Th_{t})},
\end{split}
\end{equation}
where $\theta$ controls the flatness of the distribution.
This cache distribution is then interpolated with the model's output distribution over the vocabulary. Consequently, certain words from the history are upweighted, encouraging the model to copy them. 

\paragraph{Caches help words that can be copied from long-range context the most.}
In order to study the effectiveness of the cache for the three classes of words ($C_{\text{near}}, C_\text{far}, C_{\text{none}}$), we evaluate an LSTM language model with and without a cache, and measure the difference in perplexity for these words.
In both settings, the model is provided all prior context (not just 300 tokens) in order to replicate the \citet{grave17cache} setup.
The amount of history recorded, known as the cache size, is a hyperparameter set to 500 past timesteps for PTB and 3,875 for Wiki, both values very similar to the average document lengths in the respective datasets.
\begin{figure}
	\includegraphics[width=\linewidth]{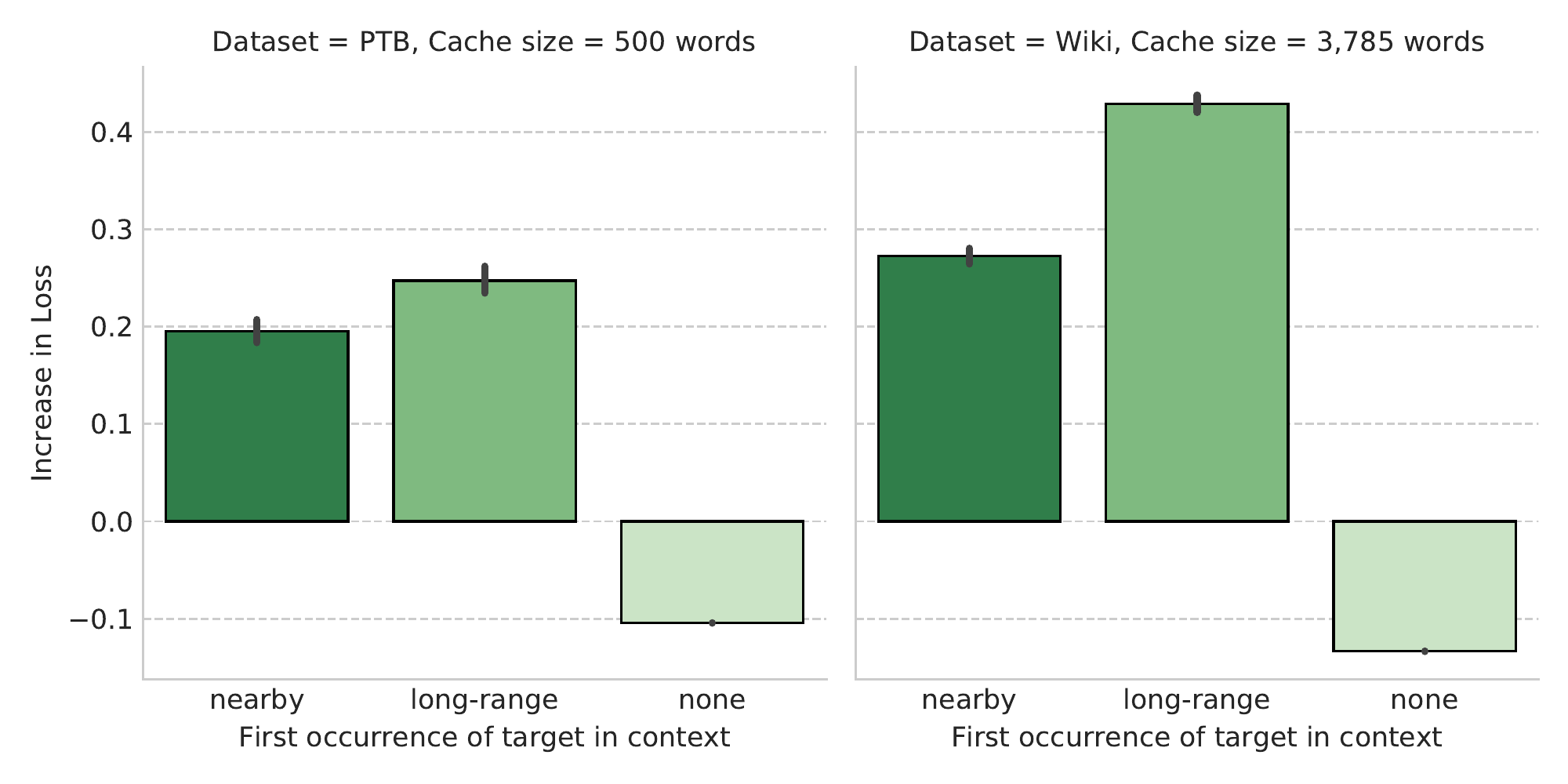}
	\caption{Model performance relative to using a cache. Error bars represent 95\% confidence intervals. Words that can only be copied from the distant context benefit the most from using a cache.
	}
	\label{fig:cachetarget}
\end{figure}

We find that the cache helps words that can only be copied from long-range context ($C_{\text{far}}$) more than words that can be copied from nearby ($C_{\text{near}}$).
This is illustrated by Figure~\ref{fig:cachetarget} where without caching, 
$C_{\text{near}}$ words see a 22\% increase in perplexity for PTB, and a 32\% increase for Wiki, whereas $C_{\text{far}}$ see a 28\% increase in perplexity for PTB, and a whopping 53\% increase for Wiki.
Thus, the cache is, in a sense, complementary to the standard model, since it especially helps regenerate words from the long-range context where the latter falls short. 

However, the cache also hurts about 36\% of the words in PTB and 20\%  in Wiki, which are words that cannot be copied from context ($C_{\text{none}}$), as illustrated by bars for ``none'' in Figure~\ref{fig:cachetarget}.
We also provide some case studies showing success (Fig.~\ref{fig:ptb_good_case}) and failure (Fig.~\ref{fig:ptb_bad_case}) modes for the cache.
We find that for the successful case, the cache distribution is concentrated on a single word that it wants to copy.
However, when the target is not present in the history, the cache distribution is more flat, illustrating the model's confusion, shown in Figure~\ref{fig:ptb_bad_case}.
This suggests that the neural cache model might benefit from having the option to ignore the cache when it cannot make a confident choice.

\section{Discussion}
\label{sec:discussion}

The findings presented in this paper provide a great deal of insight into how LSTMs model context. 
This information can prove extremely useful for improving language models. For instance, the discovery that some word types are more important than others can help refine word dropout strategies by making them adaptive to the different word types. 
Results on the cache also show that we can further improve performance by allowing the model to ignore the cache distribution when it is extremely uncertain, such as in Figure~\ref{fig:ptb_bad_case}. 
Differences in nearby vs. long-range context suggest that memory models, which feed explicit context representations to the LSTM \cite{ghosh16clstm,lau17tdlm}, could benefit from representations that specifically capture information orthogonal to that modeled by the LSTM.

In addition, the empirical methods used in this study are model-agnostic and can generalize to models other than the standard LSTM. This opens the path to generating a stronger understanding of model classes beyond test set perplexities, by comparing them across additional axes of information such as how much context they use on average, or how robust they are to shuffled contexts.

Given the empirical nature of this study and the fact that the model and data are tightly coupled, separating model behavior from language characteristics, has proved challenging. More specifically, a number of confounding factors such as vocabulary size, dataset size etc. make this separation difficult. In an attempt to address this, we have chosen PTB and Wiki - two standard language modeling datasets which are diverse in content (news vs. factual articles) and writing style, and are structured differently (eg: Wiki articles are 4-6x longer on average and contain extra information such as titles and paragraph/section markers). Making the data sources diverse in nature, has provided the opportunity to somewhat isolate effects of the model, while ensuring consistency in results. An interesting extension to further study this separation would lie in experimenting with different model classes and even different languages.

Recently, \citet{chelba2017ngram}, in proposing a new model, showed that on PTB, an LSTM language model with 13 tokens of context is similar to the infinite-context LSTM performance, with close to an 8\%
\footnote{Table 3, 91 perplexity for the 13-gram vs. 84 for the infinite context model.} 
increase in perplexity. This is compared to a 25\% increase at 13 tokens of context in our setup. We believe this difference is attributed to the fact that their model was trained with restricted context and a different error propagation scheme, while ours is not.
Further investigation would be an interesting direction for future work.

\section{Conclusion}
\label{sec:conclusion}
In this analytic study, we have empirically shown that a standard LSTM language model can effectively use about 200 tokens of context on two benchmark datasets, regardless of hyperparameter settings such as model size.
It is sensitive to word order in the nearby context, but less so in the long-range context. 
In addition, the model is able to regenerate words from nearby context, but heavily relies on caches to copy words from far away. 
These findings not only help us better understand these models but also suggest ways for improving them, as discussed in Section~\ref{sec:discussion}. 
While observations in this paper are reported at the token level, deeper understanding of sentence-level interactions warrants further investigation, which we leave to future work.

\section*{Acknowledgments}
We thank Arun Chaganty, Kevin Clark, Reid Pryzant, Yuhao Zhang and our anonymous reviewers for their thoughtful comments and suggestions.
We gratefully acknowledge support of the DARPA Communicating with Computers (CwC) program under ARO prime contract no$.$ W911NF15-1-0462 and the NSF via grant IIS-1514268.

\bibliography{acl2018}
\bibliographystyle{acl_natbib}

\clearpage
\appendix
\section{Hyperparameter settings}
\label{app:hps}
We train a vanilla LSTM language model, augmented with dropout on recurrent connections, embedding weights, and all input and output connections~\cite{wan13regularization,gal16theoretically}, weight tying between the word embedding and softmax layers~\cite{inan16tying,press16using}, variable length backpropagation sequences and the averaging SGD optimizer~\cite{merity18regopt}. We provide the key hyperparameter settings for the model in Table~\ref{tab:hps}. These are the default settings suggested by~\cite{merity18regopt}. 

\begin{table}
	\centering
	\small
	\setlength{\tabcolsep}{0.4em}
	\begin{tabular}{m{3.0cm}cccc}
		\toprule
		\textbf{Hyperparameter} & \textbf{PTB} & \textbf{Wiki}\\
		\midrule	
		Word Emb. Size & 400 & 400\\
		Hidden State Dim & 1150 & 1150\\
		Layers & 3 & 3\\
		Optimizer & ASGD & ASGD\\
		Learning Rate & 30 & 30\\
		Gradient clip & 0.25 & 0.25\\
		Epochs (train) & 500 & 750\\
		Epochs (finetune) & 500 (max) & 750 (max)\\
		Batch Size & 20 & 80\\
		Sequence Length & 70 & 70\\
		LSTM Layer Dropout & 0.25 & 0.2\\
		Recurrent Dropout & 0.5 & 0.5\\
		Word Emb. Dropout & 0.4 & 0.65\\
		Word Dropout & 0.1 & 0.1\\ 
		FF Layers Dropout & 0.4 & 0.4\\
		Weight Decay & $1.2\times10^{-6}$& $1.2\times10^{-6}$\\
		\bottomrule
	\end{tabular}
	\caption{Hyperparameter Settings.}
	\label{tab:hps}
\end{table}

\section{Additional Figures}
\label{app:figs}
This section contains all figures complementary to those presented in the main text. Some figures, such as Figures~\ref{fig:contexthps},~\ref{fig:poscontextsize} etc. present results for only one of the two datasets, and we present the results for the other dataset here. It is important to note that the analysis and conclusions remain unchanged. 
Just as before, all results are averaged from three models trained with different random seeds. Error bars on curves represent the standard deviation and those on bar charts represent 95\% confidence intervals.

\begin{figure}[h]
	\includegraphics[width=\linewidth]{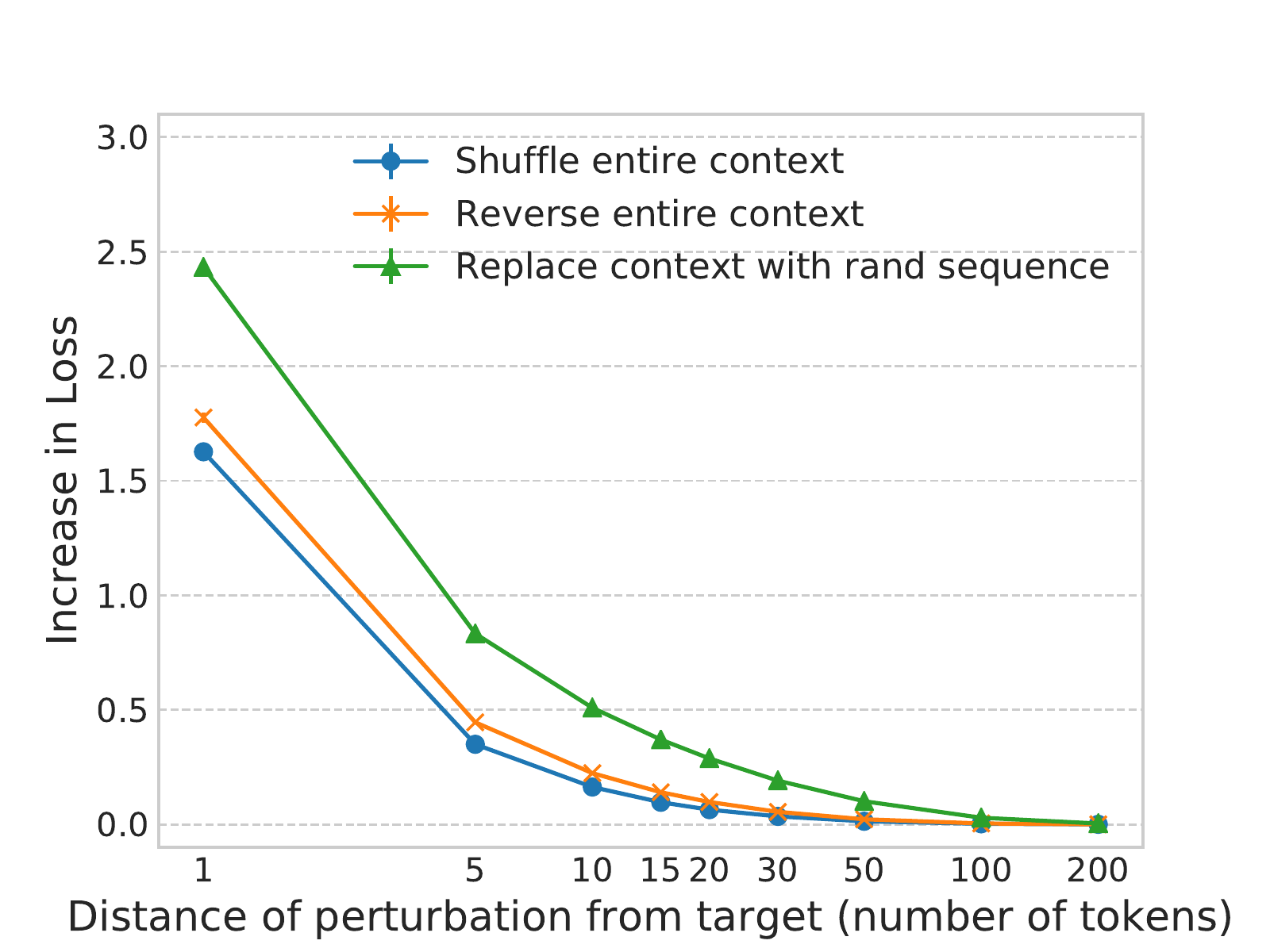}
	\caption{Complementary to Figure~\ref{fig:globalorder}. Perturb global order, i.e. all tokens in the context before a given point, in PTB. Effects of shuffling and reversing the order of words in 300 tokens of context, relative to an unperturbed baseline. Changing the global order of words within the context does not affect loss beyond 50 tokens.}
	\label{fig:ptb_global_order}
\end{figure}

\begin{figure}
	\includegraphics[width=\linewidth]{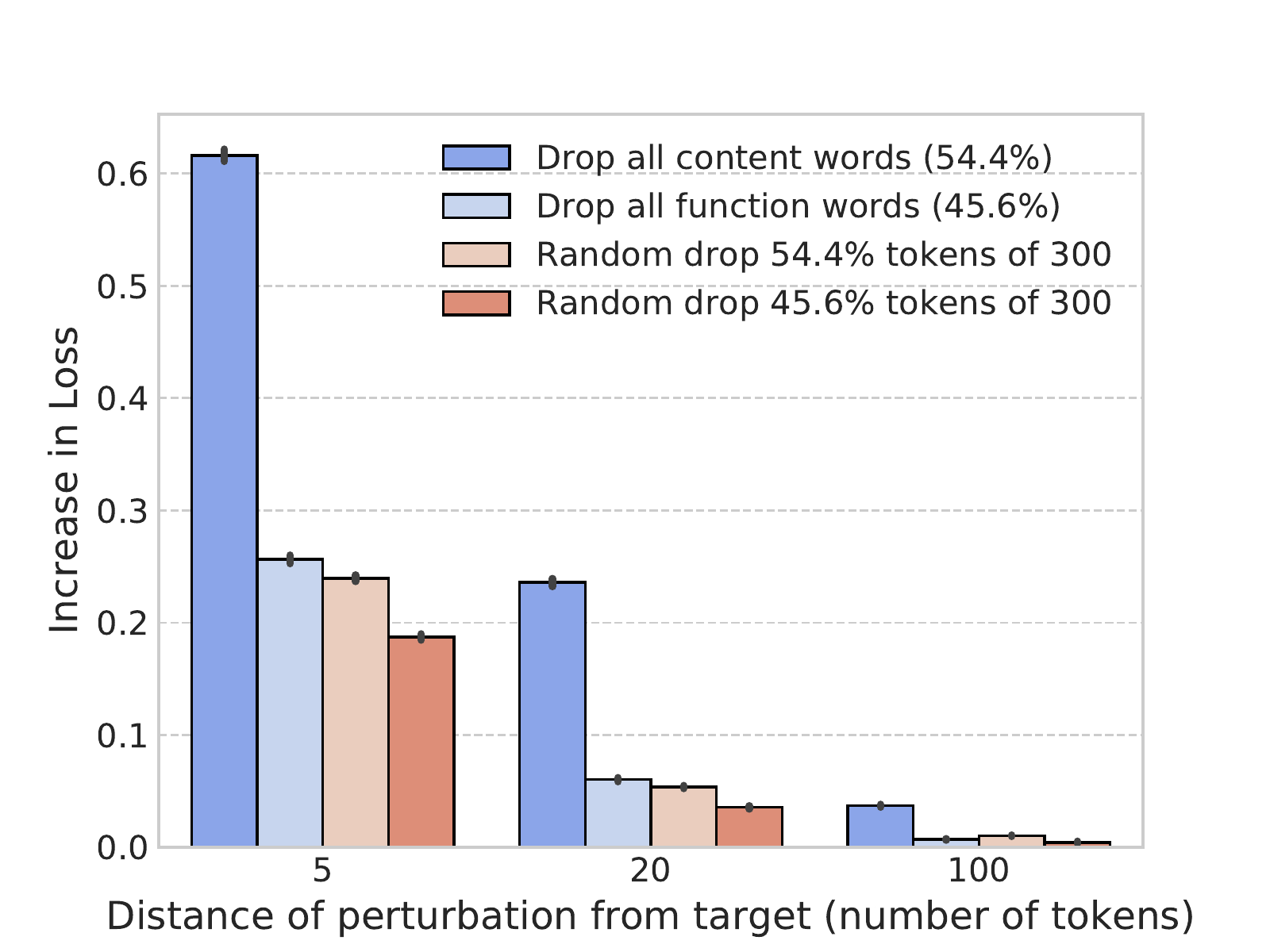}
	\caption{Complementary to Figure~\ref{fig:posdrop}. Effect of dropping content and function words from 300 tokens of context relative to an unperturbed baseline, on Wiki. Dropping both content and function words 5 tokens away from the target results in a nontrivial increase in loss, whereas beyond 20 tokens, content words are far more relevant.}
	\label{fig:wikiposdrop}
\end{figure}

\begin{figure*}
	\centering
	\begin{subfigure}[t]{0.49\textwidth}
		\centering
		\includegraphics[width=0.9\textwidth]{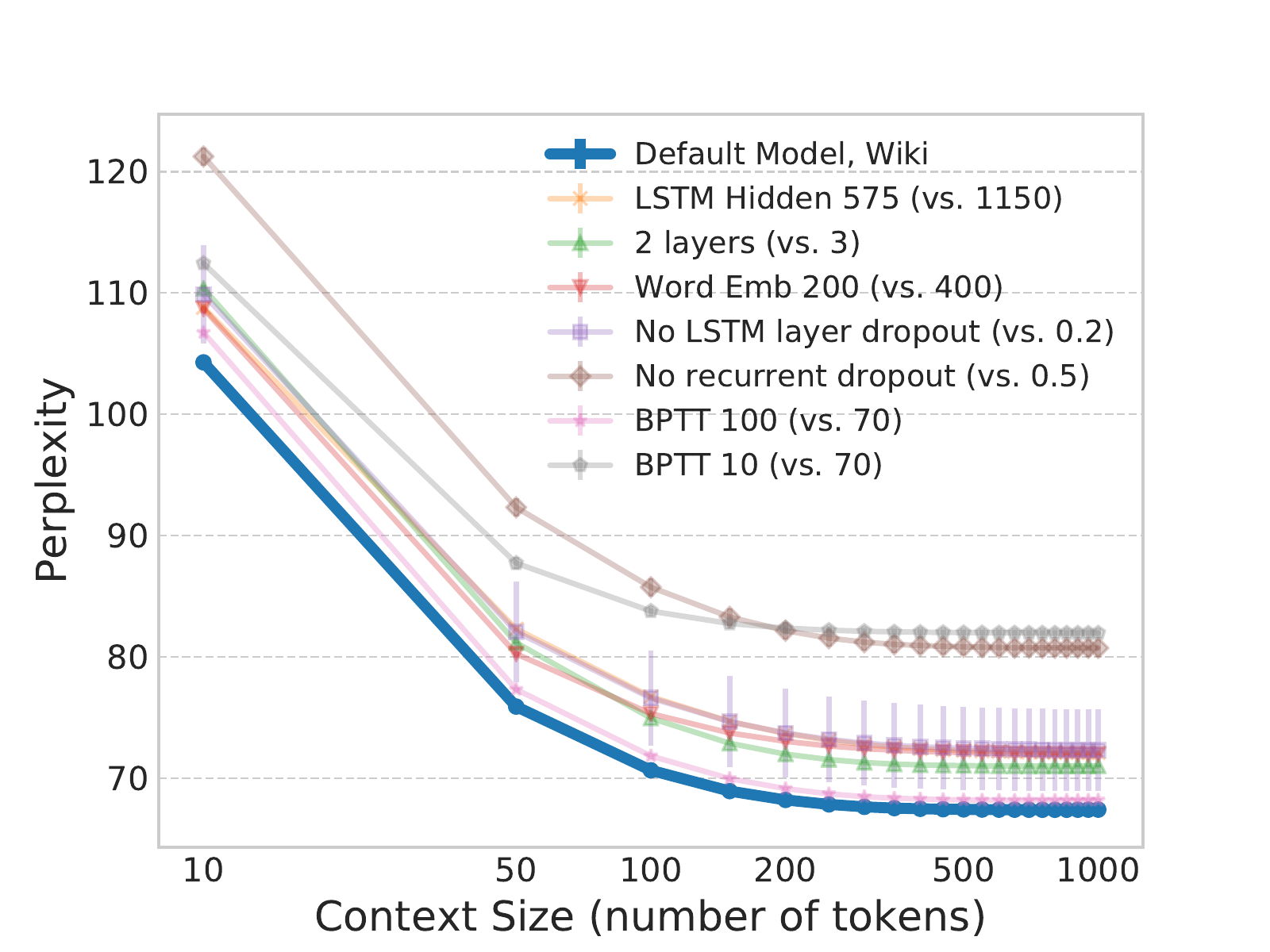}
		\caption{Changing model hyperparameters for Wiki.}
		\label{fig:wiki_context_hps}
	\end{subfigure}
	\hfill
	\begin{subfigure}[t]{0.49\textwidth}
		\centering
		\includegraphics[width=0.9\textwidth]{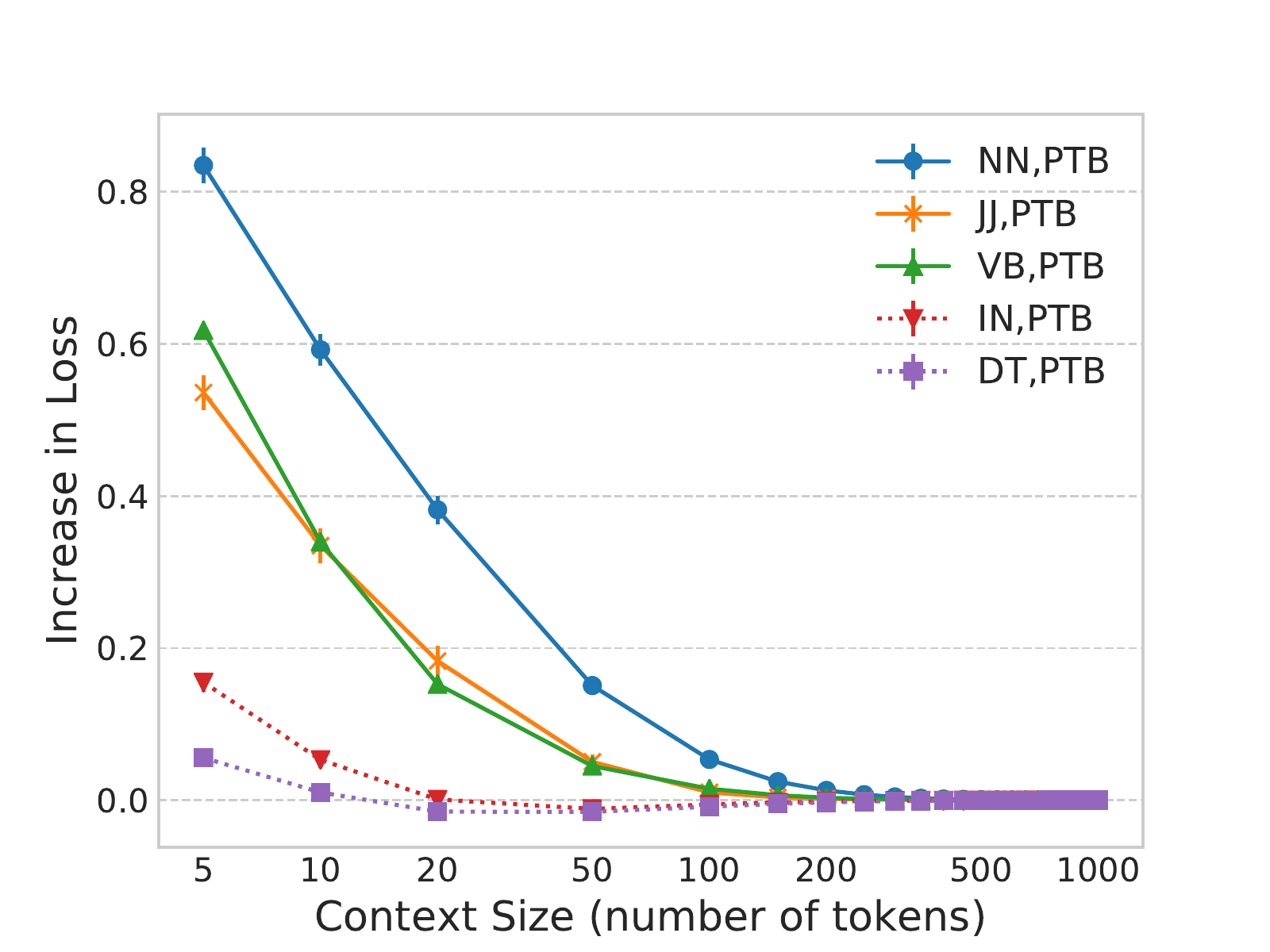}
		\caption{Different parts-of-speech for PTB.}
		\label{fig:ptb_pos_context_size}
	\end{subfigure}
	\caption{Complementary to Figures~\ref{fig:contexthps} and~\ref{fig:poscontextsize}, respectively. Effects of varying the number of tokens provided in the context, as compared to the same model provided with infinite context. Increase in loss represents an absolute increase in $\nll$ over the entire corpus, due to restricted context. \textbf{(a)} Changing model hyperparameters does not change the context usage trend, but does change model performance. We report perplexities to highlight the consistent trend. \textbf{(b)} Content words need more context than function words.}
	\label{fig:app_context_size}
\end{figure*}

%

\begin{figure*}
	\centering
	\begin{subfigure}[t]{0.49\textwidth}
		\centering
		\includegraphics[width=0.9\textwidth]{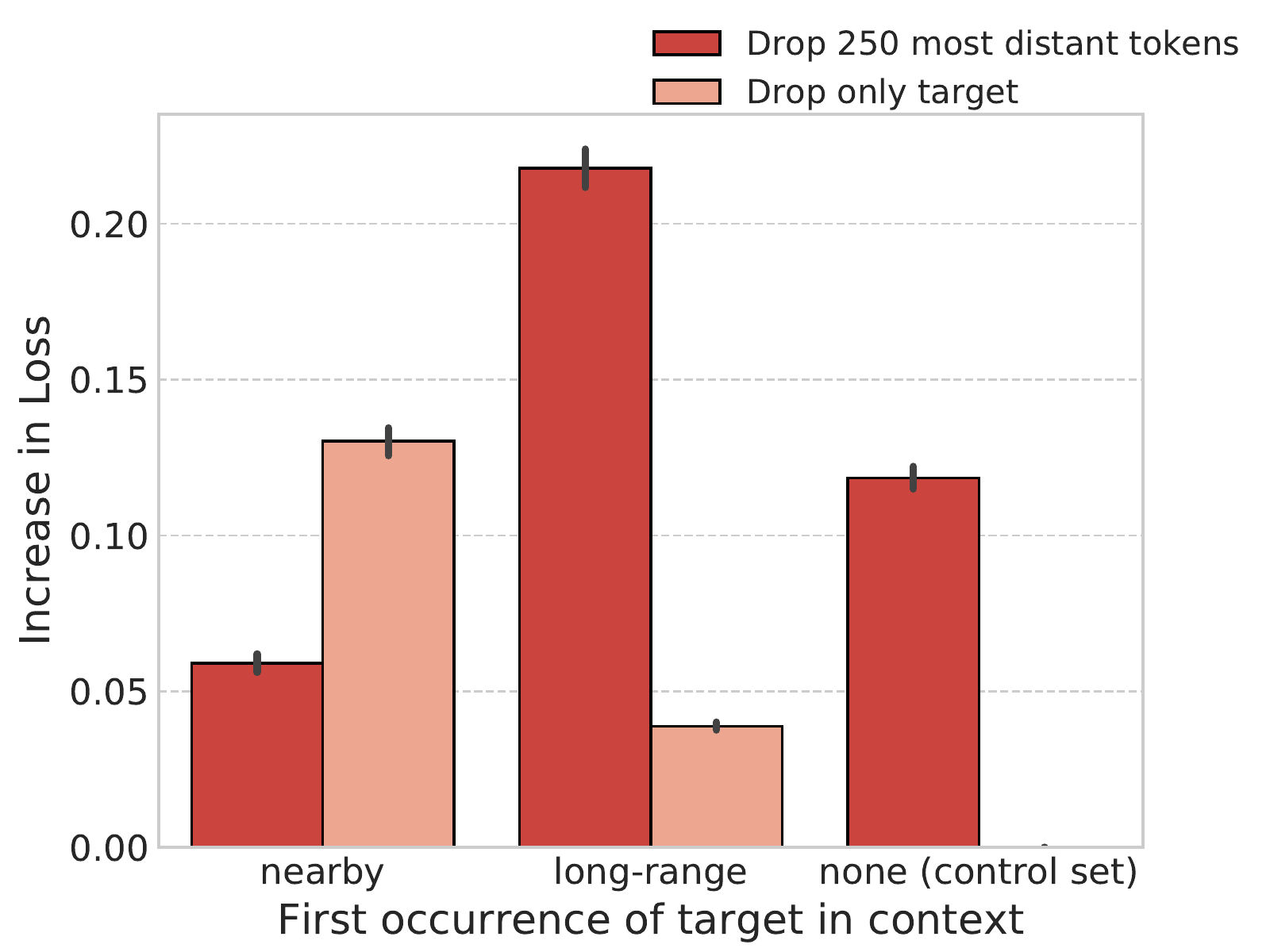}
		\caption{Dropping tokens}
		\label{fig:wiki_droptarget}
	\end{subfigure}
	\hfill
	\begin{subfigure}[t]{0.49\textwidth}
		\centering
		\includegraphics[width=0.9\textwidth]{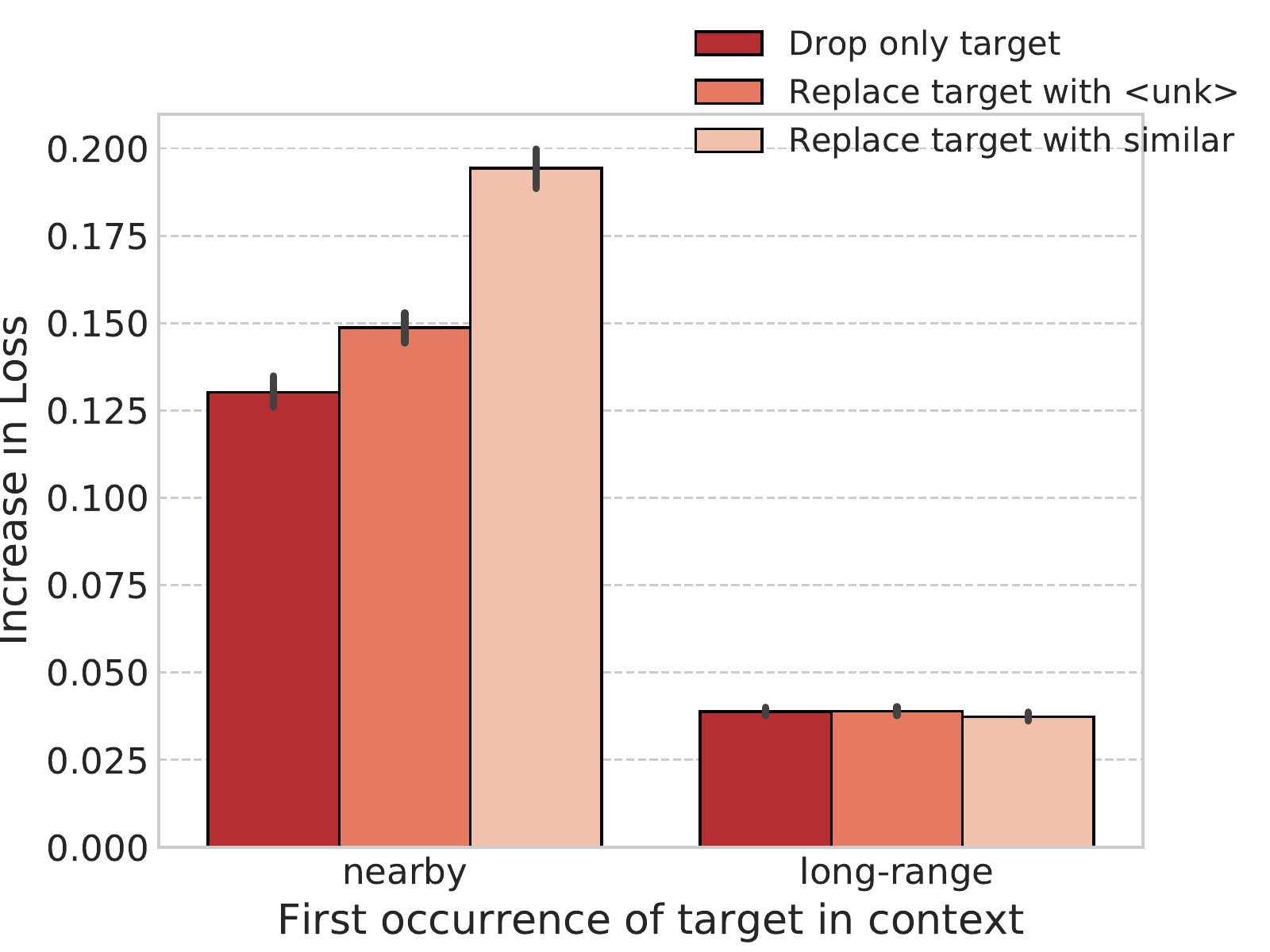}
		\caption{Perturbing occurrences of target word in context.}
		\label{fig:wiki_perturbtarget}
	\end{subfigure}
	\caption{Complementary to Figure~\ref{fig:target}. Effects of perturbing the target word in the context compared to dropping long-range context altogether, on Wiki. \textbf{(a)} Words that can only be copied from long-range context are more sensitive to dropping all the distant words than to dropping the target. For words that can be copied from nearby context, dropping only the target has a much larger effect on loss compared to dropping the long-range context. \textbf{(b)} Replacing the target word with other tokens from vocabulary hurts more than dropping it from the context, for words that can be copied from nearby context, but has no effect on words that can only be copied from far away.}
\label{fig:wiki_target}
\end{figure*}

\begin{figure*}[!b]
    \includegraphics[width=\linewidth]{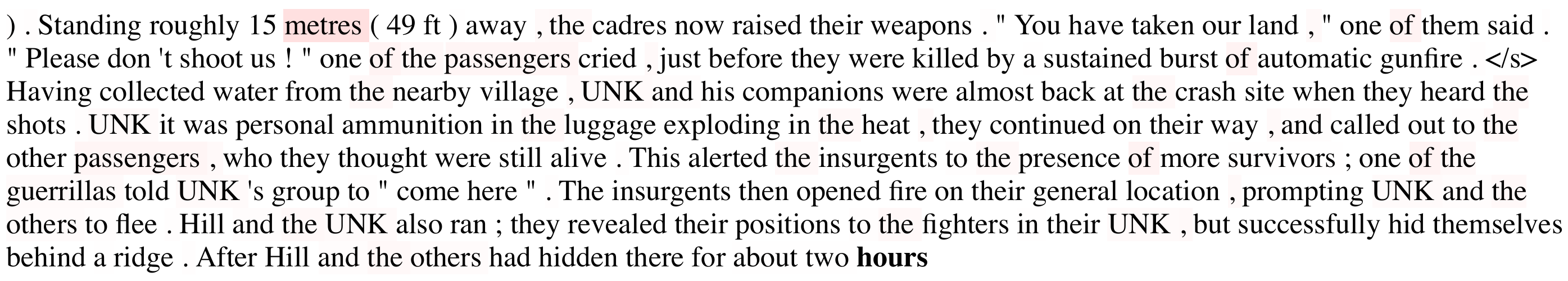}
    \caption{Failure of neural cache on Wiki. Lightly shaded regions show flat distribution.}
    \label{fig:wiki_bad_case}
\end{figure*}
\begin{figure*}[!b]
    \includegraphics[width=\linewidth]{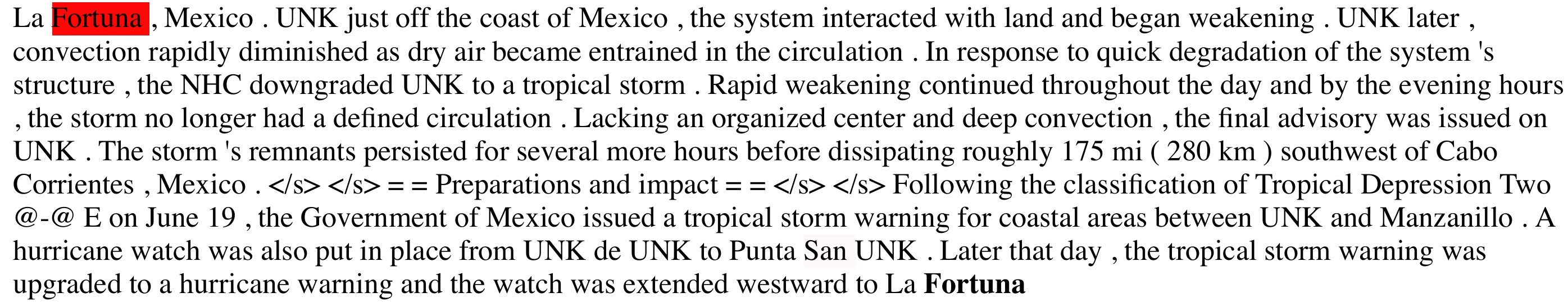}
    \caption{Success of neural cache on Wiki. Brightly shaded region shows peaky distribution.}
    \label{fig:wiki_good_case}
\end{figure*}

\end{document}